\pdfoutput=1
\PassOptionsToPackage{table,xcdraw}{xcolor}
\documentclass[11pt]{article}

\usepackage[final]{acl}

\usepackage{times}
\usepackage{latexsym}

\usepackage[T1]{fontenc}

\usepackage[utf8]{inputenc}

\usepackage{microtype}

\usepackage{inconsolata}

\usepackage{graphicx}
\usepackage[nointegrals]{wasysym}
\usepackage{subcaption}
\title{Unlocking Memorization in Large Language Models with Dynamic Soft Prompting}

\author{
 \textbf{Zhepeng Wang\textsuperscript{1}},
 \textbf{Runxue Bao\textsuperscript{2}},
 \textbf{Yawen Wu\textsuperscript{3}},
 \textbf{Jackson Taylor\textsuperscript{4}},
\\
 \textbf{Cao Xiao\textsuperscript{2}},
 \textbf{Feng Zheng\textsuperscript{5}},
 \textbf{Weiwen Jiang\textsuperscript{1}},
 \textbf{Shangqian Gao\textsuperscript{6}}\thanks{Co-corresponding authors.}, 
 \textbf{Yanfu Zhang\textsuperscript{4}}\footnotemark[1]
\\
 \textsuperscript{1}George Mason University,
 \textsuperscript{2}GE Healthcare, \\
 \textsuperscript{3}University of Pittsburgh, 
 \textsuperscript{4}William and Mary, \\
 \textsuperscript{5}Southern University of Science and Technology,
 \textsuperscript{6}Florida State University
\\
\textsuperscript{1}\texttt{\{zwang48, wjiang8\}@gmu.edu},
\textsuperscript{2}\texttt{\{runxue.bao,  cao.xiao\}@gehealthcare.com},\\
\textsuperscript{3}\texttt{yawen.wu@pitt.edu},
\textsuperscript{4}\texttt{\{jttaylor01, yzhang105\}@wm.edu},\\
\textsuperscript{5}\texttt{zfeng02@gmail.com},
\textsuperscript{6}\texttt{sgao@cs.fsu.edu}
\\
}

\usepackage{multirow}
\usepackage{xcolor}

\def\framework{our method}
\def\amazonbaseline{CSP~\citep{ozdayi2023controlling} }
\def\amazonbaselinetable{CSP~\citep{ozdayi2023controlling} }

\usepackage{bbm}
\usepackage{enumitem}
\usepackage{bm}
\usepackage{amsmath}  
\usepackage{hyperref}
\usepackage{xurl} 

\usepackage{pifont}

\begin{document}
\maketitle
\begin{abstract}

Pretrained large language models (LLMs) have revolutionized natural language processing (NLP) tasks such as summarization, question answering, and translation. However, LLMs pose significant security risks due to their tendency to memorize training data, leading to potential privacy breaches and copyright infringement. Accurate measurement of this memorization is essential to evaluate and mitigate these potential risks. However, previous attempts to characterize memorization are constrained by either using prefixes only or by prepending a constant soft prompt to the prefixes, which cannot react to changes in input. To address this challenge, we propose a novel method for estimating LLM memorization using dynamic, prefix-dependent soft prompts. Our approach involves training a transformer-based generator to produce soft prompts that adapt to changes in input, thereby enabling more accurate extraction of memorized data. Our method not only addresses the limitations of previous methods but also demonstrates superior performance in diverse experimental settings compared to state-of-the-art techniques. In particular, our method can achieve the maximum relative improvement of 112.75\% and 32.26\% over the vanilla baseline in terms of \textit{discoverable memorization rate} for the text generation task and code generation task respectively.
\end{abstract}

\section{Introduction}
\label{sec:intro}
Pretrained large language models (LLMs)~\citep{brown2020language, touvron2023llama, almazrouei2023falcon, jin2024learning} have achieved remarkable success across a wide range of downstream natural language processing (NLP) tasks such as summarization~\citep{zhang2024benchmarking,van2024adapted,zhang2019pretraining}, classification~\cite{wang2023large, sun2023text, wang2024infuserki, gao2024adaptive}, question answering~\citep{pan2023retrieving,zhang2024pruning,shao2023prompting,louis2024interpretable,jiang2021can,guo2023images,yasunaga2021qa} and translation~\citep{zhang2023prompting,bawden2023investigating,he2024exploring,xue2020mt5,xu2023paradigm,li2024pre}, etc. The popularity of LLMs requires people to pay attention to the unique challenges they bring to security. One of the significant security issues is that LLMs can memorize a considerable portion of their training data even though they tend to not overfit to their training dataset due to the small number of training epochs~\citep{radford2019better}. Moreover, the memorized data can be extracted by carefully designed input from attackers or even unintentional input from ordinary users, which can cause privacy and copyright issues with the sensitive training data~\citep{carlini2021extracting, carlini2023quantifying, ozdayi2023controlling, nasr2023scalable, karamolegkou2023copyright}. For example, the confidential codes from Samsung can be exposed to other users after they were shared with OpenAI due to the memorization of LLMs~\citep{Samsungleakage, Starcoderleakage}.
\begin{figure}[!t]
\centering
\includegraphics[width=0.45\textwidth]{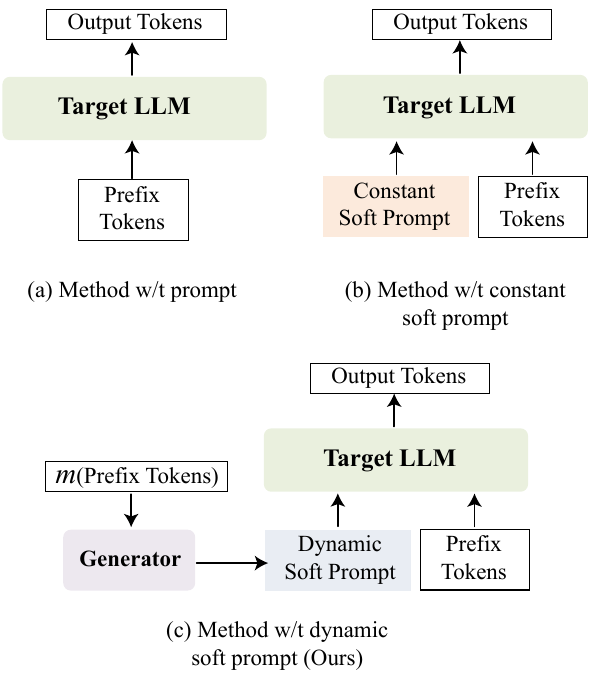}
\caption{Conceptual comparison of three methods for extracting memorized data from the target LLM.}

\label{fig:motivation}
\end{figure}

The huge security risks and the potential uses of memorization make it important to measure the memorization of the target LLM. With an accurate method to quantify the intrinsic memorization of LLMs, model developers can have a better understanding of the model's vulnerability posed by its memorization and take actions such as machine unlearning~\citep{yao2023large,pawelczyk2023context, yao2024machine} to mitigate the memorization before they release their LLMs to the public. Moreover, the method to extract memorized data can also be combined with the target LLM and leveraged by the users to detect whether their self-built dataset has data leakage issues when it is used to evaluate the target LLM.

To measure the intrinsic memorization of the target LLM,~\citet{carlini2023quantifying} first proposed a metric called \textit{discoverable memorization rate} to serve as the estimation. As shown in Figure~\ref{fig:motivation} (a), the given data are split into prefix tokens and suffix tokens and the prefix tokens are fed into the target LLM. The data are defined as discoverably memorized when the output tokens can match the suffix tokens verbatim.~\citet{ozdayi2023controlling} proposed that more memorized data can be extracted via learning a soft prompt and prepending it to the prefix tokens for generation, which is shown in Figure~\ref{fig:motivation} (b). However, the memorization of the target LLM is still underestimated even with the soft prompt since it is constant and invariant to prefix tokens, which may not help or even hinder extracting data when changing the prefix.

In this paper, we propose a new method to estimate the memorization of LLMs. Compared with constant soft prompts~\citep{ozdayi2023controlling}, our method can generate prefix-dependent soft prompts and react to the changes in inputs. More specifically, a transformer-based generator is trained for the generation of the dynamic soft prompt. As shown in Figure~\ref{fig:motivation} (c), it takes the outputs from the naive mapping $m(\cdot)$ of prefix tokens as its input and emits the corresponding soft prompt prepended to the prefix tokens. This method can customize prompts given the inputs and thus extract more memorized data from the target LLM, which can reflect its intrinsic memorization more accurately.

Our contributions can be summarized as follows.
\begin{itemize}[leftmargin=*]
\item We propose a new method with dynamic soft prompts to extract memorized data from the target LLMs and estimate their memorization with the same assumption as the state-of-the-art (SOTA) work~\citep{ozdayi2023controlling} but overcoming its limitation on the invariance to the input.
\item We develop a transformer-based generator to produce the dynamic soft prompts in response to the change of input. To find the best parameters of the generator, we utilize a technique to initialize the transformer blocks within the generator as identity mappings for the effective and robust training of the generator.

\item We evaluate~\framework~on more diverse settings than that of the SOTA work~\citep{ozdayi2023controlling}. Experimental results show that~\framework~can outperform all the baselines consistently in all the evaluated settings. The maximum relative improvement of 112.75\% and 32.26\% are achieved over the vanilla baseline for the text generation and code generation tasks, respectively.

\end{itemize}
\section{Related Work}
\label{sec:related work}

\noindent\textbf{LLM Memorization.} The memorization of LLM is firstly verified by \citet{carlini2021extracting}. It shows that it is feasible for attackers to extract training data from target LLMs by producing a large number of random prefixes and feeding them to the target LLM for generation.~\citet{carlini2023quantifying} then defines the concept of \textit{discoverably memorized} and utilizes it to quantify the memorization of the target LLM. In addition to the memorization of pretrained LLM on the pretraining dataset,~\citet{zeng2023exploring} studies the memorization of fine-tuned LLM on the fine-tuning dataset. It shows that memorization also exists in fine-tuning settings and that the characteristics of memorization vary with the type of fine-tuning tasks.~\citet{karamolegkou2023copyright} shows that the memorization of LLM can cause copyright violations for books and proprietary codes.~\citet{nasr2023scalable} demonstrates that it is feasible to extract gigabytes of training data from production LLMs such as ChatGPT due to their memorization. Recently, ~\citet{ozdayi2023controlling} proposes to learn a constant soft prompt to extract more training data from LLM to measure memorization. However, we argue that this method still underestimates the memorization of LLM since the soft prompt is independent of the input and thus does not react to the dynamics of the input. Our method can address these limitations.\\
\noindent\textbf{Defend against Memorization.} Training LLMs with differentially private training~\citep{abadi2016deep} is considered effective in preventing the memorization of individual training samples with a theoretical guarantee~\citep{carlini2021extracting}, However, the training cost is expensive --- even prohibitive for LLMs. Moreover, the utility of LLMs is significantly degraded, making them impractical for real-world applications. Alternatively, deduplicating training data can mitigate LLM memorization~\citep{lee2021deduplicating, kandpal2022deduplicating}. However, it cannot eliminate the memorization since certain portions of data will be memorized by LLM inevitably even if they only appear once in the training data. Similarly,~\citet{ippolito2023preventing} shows that memorization can not be prevented by applying runtime filtering to the user input. Therefore, the ``ultimate'' solution to prevent memorization is still under exploration. Machine unlearning~\citep{yao2023large,pawelczyk2023context, yao2024machine} is a promising method to defend against memorization. By identifying the set of memorized training data to be the forget set for unlearning, LLM can forget these data via gradient ascent~\citep{yao2023large} or in-context learning~\citep{pawelczyk2023context}. Compared to existing methods, our method can identify a larger and more accurate forget set for machine unlearning to defend against memorization.\\
\textbf{Prompt Tuning.} Training or finetuning machine learning models is usually costly. To enable efficient training, a variety of methods are proposed to reduce the training cost via pruning~\citep{bao2020fast, bao2022accelerated, bao2022doubly}, data selection~\citep{shrivastava2016training, wang2019e2, wu2021enabling} or parameter selection~\citep{wu2020enabling, hu2021lora, liu2022few,wang2024infuserki}, etc. All of these methods require adapting the internal parameters of the target model. Therefore, applying these methods to finetuning the LLM may still be expensive due to the large number of parameters of the LLM. Prompt tuning, introduced by~\citet{lester2021power}, is an efficient method for adapting pre-trained models to various tasks by learning "soft prompts" that condition frozen language models without changing their internal parameters.  In the realm of NLP, researchers have harnessed trainable representations in the form of soft prompts using methods like prompt-tuning, with~\citet{su2022transferability} and~\citet{vu2022spot} demonstrating successful transferability and improved performance. \citet{ma2022xprompt} uses pruning to remove ineffective tokens, and~\citet{wei2021pretrained} provides theoretical proof of prompt tuning's downstream guarantees under weaker non-degeneracy conditions. Prompt tuning has also been applied to vision tasks~\citep{jia2022visual, lian2022scaling, chen2022adaptformer}, including continual learning~\citep{wang2022learning} and image inpainting~\citep{bar2022visual}. Different from previous work that used prompt tuning to improve downstream performance, our work leverages continuous prompts to more accurately reflect intrinsic memorization, extract memorized data from the target LLMs, and measure their memorization.
\section{Method}
\label{sec:method}
\begin{figure*}[!t]
\centering
\includegraphics[width=0.8\textwidth]{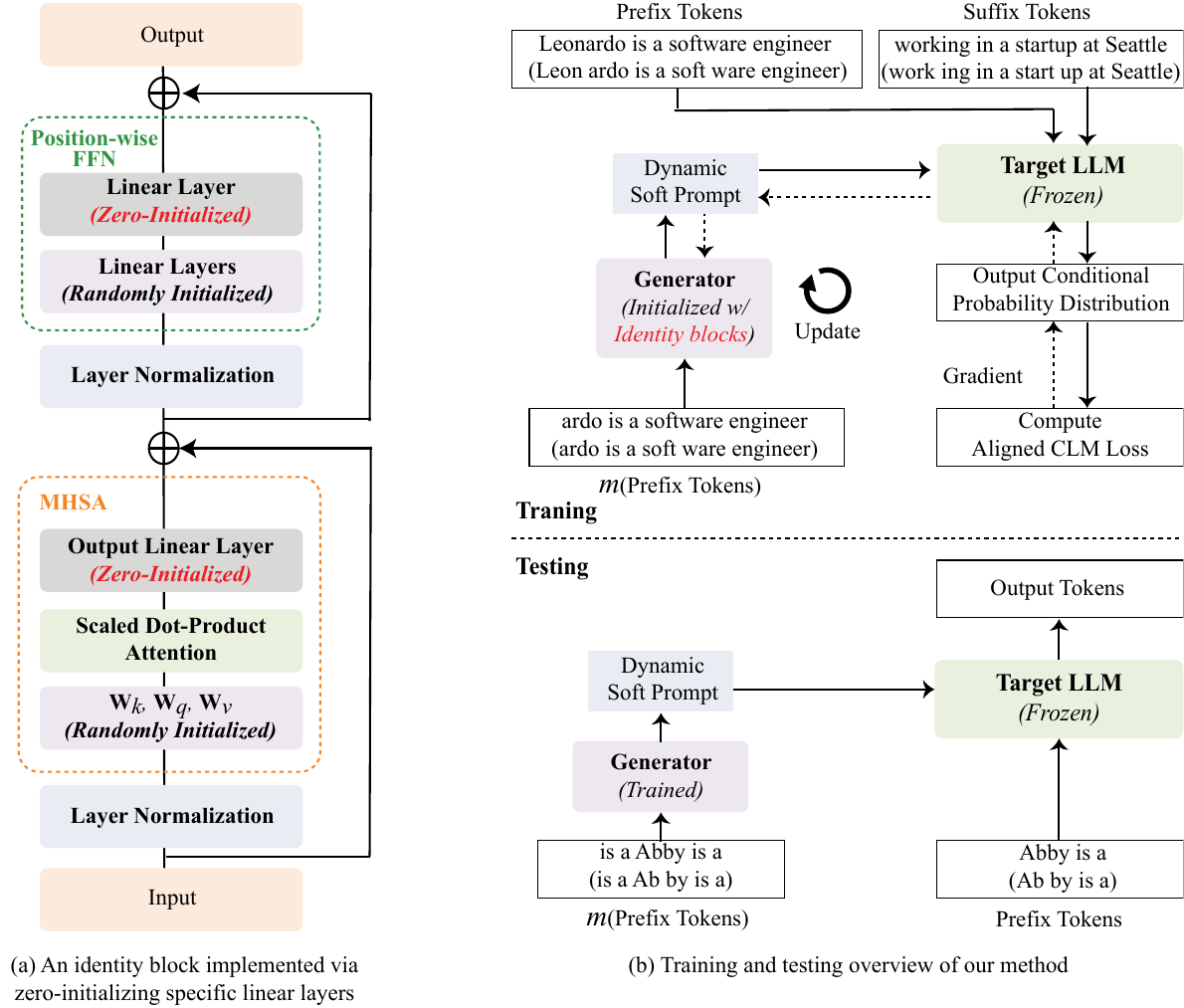}
\caption{Illustration of our method.}
\label{fig:overview}
\end{figure*}

\subsection{Problem Formulation}
\label{subsec:problem formulation}
According to the work ~\citep{nasr2023scalable}, given the target LLM $f_{\bm{\theta}}$ and data $x$, $x$ is defined as \textit{discoverably memorized} if there exists a generation routine $G$, such that $f_{\bm{\theta}}(G(p)) = s$, where $x = [p || s]$ and $x$ is split into prefix $p$ and suffix $s$. The generation routine can be constant soft prompts~\citep{ozdayi2023controlling}, dynamic soft prompts (our method), or just the identity function~\citep{carlini2023quantifying}.

In our problem setting, a set of sequences $\mathcal{D}_{\text{tr}}$ is randomly sampled from the training set $\mathcal{D}$ of the target LLM $f_{\bm{\theta}}$, we aim to find the generation routine $G$ to maximize \textit{discoverable memorization rate} over the training set $\mathcal{D}$ by leveraging $\mathcal{D}_{\text{tr}}$. We use another disjoint set $\mathcal{D}_{\text{test}}$ randomly sampled from $\mathcal{D}$ to evaluate the \textit{discoverable memorization rate} over $\mathcal{D}$, which is defined as,

\begin{equation}
\label{eq:problem_def}
    \max \frac{1}{|\mathcal{D}_{\text{test}}|}\sum_{x_{i} \in \mathcal{D}_{\text{test}}} \mathbbm{1}_{f_{\bm{\theta}}(G( p_{i}))=s_{i}}(p_{i})
\end{equation}
where $\mathbbm{1}(\cdot)$ denotes the indicator function and $x_{i} = [p_{i} || s_{i}]$.

\subsection{Method Overview}
To maximize the \textit{discoverable memorization rate}, we propose a pipeline to learn a transformer-based generator $g_{\bm{\omega}}$ to build the generation routine $G$. As shown in Figure~\ref{fig:overview} (b), the generator $g_{\bm{\omega}}$ is initialized with $K$ identity blocks, which are illustrated in Section~\ref{subsec:identity}. The input to $g_{\bm{\omega}}$ is $m(p)$, where $m(\cdot)$ represents a naive mapping of prefix tokens $p$ and it is detailed in Section~\ref{subsec:mapping}. The dynamic soft prompt $o$ is then generated via $g_{\bm{\omega}}$, where $o = g_{\bm{\omega}}(m(p))$. Since $o$ depends on the prefix token $p$, it can adapt to the change in $p$. Note that the dimension of $o$ should be the same as the dimension of the embedding $E(x)$ of the target LLM $f_{\bm{\theta}}$ for its concatenation with the input data $x$.

We train the generator $g_{\bm{\omega}}$ on $\mathcal{D}_{\text{tr}}$ to obtain the optimized parameters $\bm{\omega}^{*}$. For each sequence $x_{i} \in \mathcal{D}_{\text{tr}}$, where $x_{i} = [p_{i} || s_{i}]$, the dynamic soft prompt $o_{i}$ is generated and then prepended to the embeddings $E(p_{i})$ of prefix tokens $p_{i}$ and the embeddings $E(s_{i})$ of suffix tokens $s_{i}$. Thus, we obtain the input $q_{i}$ to the target LLM $f_{\bm{\theta}}$, where $q_{i} =[o_{i}||E(p_{i})||E(s_{i})]$. By feeding $q_{i}$ to the target LLM $f_{\bm{\theta}}$, we aim to minimize the aligned causal language modeling (CLM) loss $\mathcal{L}$~\citep{ozdayi2023controlling} over $\mathcal{D}_{\text{tr}}$, which is defined as,
\begin{equation}
\label{eq:loss}
\mathcal{L} = -\sum_{x_{i} \in \mathcal{D}_{\text{tr}}}\sum_{t=k_{i}}^{|q_{i}|-1}\log P_{\bm{\theta},\bm{\omega}}(q_{i, t}|q_{i, 1}, ..., q_{i, t-1}),
\end{equation}
where $q_{i, t}$ represents the $t$ th token in the input sequence $q_{i}$. $P_{\bm{\theta},\bm{\omega}}(q_{i, t}|q_{i, 1}, ..., q_{i, t-1})$ denotes the  output conditional probability distribution at the $t$ th token given the preceding $t-1$ tokens. $k_{i}$ represents the index of the starting token in suffix $s_{i}$. Therefore, the aligned CLM loss only focuses on the token prediction at the position of suffix tokens, which aligns with the definition of \textit{discoverable memorization}. During the training phase, only the parameters $\bm{\omega}$ of $g_{\bm{\omega}}$ are updated based on the gradients calculated from the aligned CLM loss while the parameters $\bm{\theta}$ of $f_{\bm{\theta}}$ are frozen.

During the testing phase of the trained generator $g_{\bm{\omega}^{*}}$, for each testing sequence $x_{i} \in \mathcal{D}_{test}$,  only the dynamic soft prompt $o_{i}$ and the embedding of prefix tokens $E(p_{i})$ are concatenated and sent to the target LLM $f_{\bm{\theta}}$ for generation. The generated output tokens $y_{i}$ are then compared with the suffix tokens $s_{i}$ for evaluation, where $y_{i} = f_{\bm{\theta}}([o_{i}||E(p_{i})])$.

\subsection{Mapping of Prefix Tokens}
\label{subsec:mapping}
\begin{figure}[!t]
\centering
\includegraphics[width=0.45\textwidth]{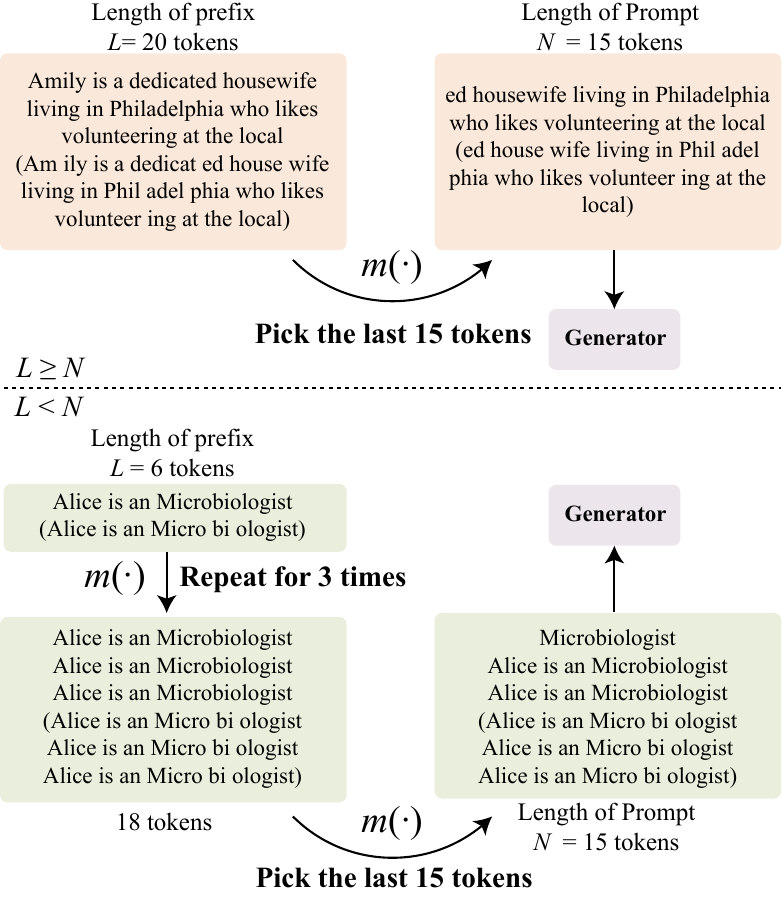}
\caption{Illustration of naive mapping $m(\cdot)$ with examples.}
\vspace{-15pt}
\label{fig:mapping}
\end{figure}
According to the constant soft prompt~\citep{ozdayi2023controlling}, the length of the prompt $N$ is a hyperparameter of the method and its value can affect the extraction of data. If we feed the prefix tokens $p$ to $g_{\bm{\omega}}$ directly, then the length of the dynamic soft prompt $o$ will be limited to the length of the prefix tokens $p$. To provide the same flexibility as the constant soft prompt~\citep{ozdayi2023controlling}, we propose a naive mapping $m(\cdot)$ to preprocess the prefix tokens $p$ and send its output $m(p)$ to the generator $g_{\bm{\omega}}$.

The details of $m(\cdot)$ are shown in Figure~\ref{fig:mapping} with an example. Assume the length of $p$ and $m(p)$ is $L$ and $N$, respectively. If $L\geq N$, $m(p)$ is the last $N$ tokens of $p$. Otherwise, we first generate $r$ by duplicating $p$ for $\lceil \frac{L}{N} \rceil$ times. $m(p)$ is then the last $N$ tokens of $r$. The dynamic soft prompt $o$ is generated, where $o = g_{\bm{\omega}}(m(p))$. In this way, the length of the prompt $N$ can be an arbitrary integer, which provides the maximum flexibility for usage.

\subsection{Identity Blocks with Zero-Initialization}
\label{subsec:identity}

\begin{table}[!t]
\centering
\small
\caption{Case Study on the Effect of Identity Blocks with Zero-Initialization}
\label{tab:zero-init}
\resizebox{0.48\textwidth}{!}{
\begin{tabular}{ccccc}
\hline
Model &
  \begin{tabular}[c]{@{}c@{}}Is Zero- \\ Initialization?\end{tabular} &
  \begin{tabular}[c]{@{}c@{}}Is Dynamic \\ Prompt?\end{tabular} &
  \begin{tabular}[c]{@{}c@{}}Exact \\ ER\end{tabular} &
  \begin{tabular}[c]{@{}c@{}}Fractional \\ ER\end{tabular} \\ \hline
 &
  \ding{55} &
  \ding{51} &
  \textbf{0.000} &
  \textbf{0.053} \\
\multirow{-2}{*}{\textbf{\begin{tabular}[c]{@{}c@{}}GPT-Neo \\ (125M)\end{tabular}}} &
  \ding{51} &
  \ding{51} &
  0.421 &
  0.557 \\ \hline
 &
  \ding{55} &
  \ding{51} &
  \textbf{0.000} &
  \textbf{0.035} \\
\multirow{-2}{*}{\textbf{\begin{tabular}[c]{@{}c@{}}GPT-Neo \\ (1.3B)\end{tabular}}} &
  \ding{51} &
  \ding{51} &
  {\color[HTML]{343434} 0.651} &
  0.772 \\ \hline
 &
  {\color[HTML]{333333} \textbf{\ding{55}}} &
  \ding{51} &
  {\color[HTML]{343434} \textbf{0.000}} &
  \textbf{0.022} \\
\multirow{-2}{*}{\textbf{\begin{tabular}[c]{@{}c@{}}GPT-Neo \\ (2.7B)\end{tabular}}} &
  {\color[HTML]{333333} \ding{51}} &
  \ding{51} &
  {\color[HTML]{333333} 0.702} &
  0.820 \\ \hline
\end{tabular}
}
\end{table}

Randomly initializing the transformer-based generator $g_{\bm{\omega}}$ and training it from scratch may degrade its performance and even lead to model collapse. It can be verified by a case study for GPT-Neo~\citep{black_2021_5297715}, shown in Table~\ref{tab:zero-init}, where the rows without zero initialization correspond to random initializing $g_{\bm{\omega}}$. Table~\ref{tab:zero-init} shows that the random initialization performs badly with the two standard metrics for memorization being close to $0$.

The issue of random initialization is caused by the fact that the underlying latent space of the dynamic soft prompt is far away from the embedding space of the target LLM $f_{\bm{\theta}}$ at the initial stage, making it difficult for the target LLM $f_{\bm{\theta}}$ to extract meaningful information from the prompt and thus hinder the training of the generator $g_{\bm{\omega}}$. Therefore, to enable the effective and robust training of $g_{\bm{\omega}}$, it is important to align the dynamic soft prompt with the embedding of input data, making their underlying latent space close to each other. To achieve this, the tokenizer and embedding layer of the generator $g_{\bm{\omega}}$ should be initialized with those of the target LLM $f_{\bm{\theta}}$. However, this is insufficient due to the perturbation incurred by the non-identical forward pass of the transformer blocks within the generator $g_{\bm{\omega}}$. More specifically, the forward pass for each attention block can be formulated as,
\begin{subequations}
\label{eq:att_overview}
\begin{align}
z &= x + \text{MHSA}(\text{LN}(x)),\\
y &= z + \text{FFN}(\text{LN}(z)),
\end{align}
\end{subequations}
where $x$ and $y$ are the input and output of the transformer block, respectively. $z$ is the output of the attention layer within the block. MHSA($\cdot$) denotes the multi-head self-attention (MHSA) mechanism. LN($\cdot$) represents layer normalization. FFN($\cdot$) corresponds to the position-wise feed-forward network (FFN). Therefore, if the transformer block is randomly initialized, it corresponds to a non-identical function where $y \neq x$ and thus enlarges the distance between the latent space of the dynamic soft prompt and the target LLM $f_{\bm{\theta}}$.

Inspired by LLAMA PRO~\citep{wu2024llama}, we propose to initialize the transformer blocks within $g_{\bm{\omega}}$ as identity blocks to align the dynamic soft embedding with the token embedding of the target LLM $f_{\bm{\theta}}$. To illustrate the implementation of the identity block shown in Figure~\ref{fig:overview} (a), we need to delve into the details of MHSA($\cdot$) and FFN($\cdot$), which can be formulated as,

\begin{subequations}
\label{eq:att_detail}
\begin{align}
    &\text{MHSA}(x^{\prime}) = \sum_{i=1}^H\sigma_{\text{s}}(
    x^{\prime} W_{Q,i} W_{K,i}^\top x^{\prime\top}
    )x^{\prime}W_{V,i}{W_{O,i}}\label{block: mha},\\
    &\text{FFN}(z^{\prime}) = (\sigma(z^{\prime}\bm{W}_{1})\odot (z^{\prime}\bm{W}_{2})) \bm{W}_{3}\label{block: mlp},
\end{align}
\end{subequations}

where $x^{\prime}$ and $z^{\prime}$ are obtained by applying layer normalization to $x$ and $z$, respectively. Assume there are $H$ heads in MHSA($\cdot$). $W_{Q,i}$, $W_{K,i}$ and $W_{V,i}$ are the query, key and value matrix of the $i$ th head. $W_{O,i}$ is the $i$ th weight matrix of the output linear layer in the attention block. $\sigma_{\text{s}}$ denotes the softmax function. For FFN($\cdot$), $\bm{W}_{1}$ and $\bm{W}_{2}$ are the weight matrices of the first layer of linear layers within the position-wise FFN and $\sigma(\cdot)$ is the activation function, while $\bm{W}_{3}$ is the weight matrix of the second layer of the linear layer. The FFN defined in Equation~\ref{block: mlp} is regularly used in LLaMA models~\cite{touvron2023llama}. For Pythia~\cite{biderman2023pythia} or GPT-Neo~\cite{black_2021_5297715}, we have $\text{FFN}(z^{\prime}) = \sigma(z^{\prime}\bm{W}_{1}) \bm{W}_{2}$.

According to Equation~\ref{eq:att_overview} and~\ref{eq:att_detail}, we can conclude that the identity block can be built by initializing $\bm{W}_{O}$ and $\bm{W}_{3}$ as zero matrices, such that $y=x$. Moreover, it has been shown that such kind of zero initialization does not introduce zero gradients and thus does not prevent the effective training of the generator $g_{\bm{\omega}}$~\citep{wu2024llama}.

\begin{table*}[!htb]
\small
\centering
\caption{Main Results on GPT-Neo Suite
}
\label{tab:main-gptneo}
\resizebox{\textwidth}{!}{\begin{tabular}{ccccccccc}
\hline
Model &
  Method &
  \begin{tabular}[c]{@{}c@{}}Dynamic\\ Prompt?\end{tabular} &
  \begin{tabular}[c]{@{}c@{}}Exact\\ ER\end{tabular} &
  \begin{tabular}[c]{@{}c@{}}Exact ER\\ Gain\end{tabular} &
  \begin{tabular}[c]{@{}c@{}}Fractional\\ ER\end{tabular} &
  \begin{tabular}[c]{@{}c@{}}Fractional ER\\ Gain\end{tabular} &
  Test Loss &
  \begin{tabular}[c]{@{}c@{}}Test Perplexity\\ (PPL)\end{tabular} \\ \hline
 &
  No Prompt &
  N/A &
  0.189 &
  N/A &
  0.369 &
  N/A &
  0.953 &
  2.594 \\
 &
  Constant Hard Prompt &
  \ding{55} &
  0.144 &
  -23.81\% &
  0.326 &
  -11.65\% &
  1.002 &
  2.725 \\
 &
  Dynamic Hard Prompt &
  \ding{51} &
  0.056 &
  -70.37\% &
  0.153 &
  -58.54\% &
  1.122 &
  3.071 \\
 &
  \amazonbaselinetable &
  \ding{55} &
  0.239 &
  26.46\% &
  0.421 &
  14.09\% &
  0.858 &
  2.359 \\
\multirow{-5}{*}{\textbf{\begin{tabular}[c]{@{}c@{}}GPT-Neo\\ (125M)\end{tabular}}} &
  {\color[HTML]{333333} \textbf{Ours}} &
  \ding{51} &
  {\color[HTML]{333333} \textbf{0.421}} &
  \textbf{122.75\%} &
  {\color[HTML]{333333} \textbf{0.557}} &
  \textbf{50.89\%} &
  {\color[HTML]{333333} \textbf{0.665}} &
  {\color[HTML]{333333} \textbf{1.945}} \\ \hline
 &
  No Prompt &
  N/A &
  0.46 &
  N/A &
  0.643 &
  N/A &
  0.202 &
  1.224 \\
 &
  Constant Hard Prompt &
  \ding{55} &
  0.392 &
  -14.78\% &
  0.581 &
  -9.64\% &
  0.24 &
  1.271 \\
 &
  Dynamic Hard Prompt &
  \ding{51} &
  0.1 &
  -78.26\% &
  0.194 &
  -69.83\% &
  0.394 &
  1.483 \\
 &
  \amazonbaselinetable &
  \ding{55} &
  0.532 &
  15.65\% &
  0.698 &
  8.55\% &
  0.133 &
  1.142 \\
\multirow{-5}{*}{\textbf{\begin{tabular}[c]{@{}c@{}}GPT-Neo\\ (1.3B)\end{tabular}}} &
  {\color[HTML]{333333} \textbf{Ours}} &
  \ding{51} &
  {\color[HTML]{000000} \textbf{0.651}} &
  \textbf{41.52\%} &
  {\color[HTML]{000000} \textbf{0.772}} &
  \textbf{20.04\%} &
  {\color[HTML]{000000} \textbf{0.114}} &
  {\color[HTML]{000000} \textbf{1.121}} \\ \hline
 &
  No Prompt &
  N/A &
  0.54 &
  N/A &
  0.702 &
  N/A &
  0.127 &
  1.135 \\
 &
  Constant Hard Prompt &
  \ding{55} &
  0.473 &
  -12.41\% &
  0.651 &
  -7.26\% &
  0.158 &
  1.171 \\
 &
  Dynamic Hard Prompt &
  \ding{51} &
  0.117 &
  -78.33\% &
  0.213 &
  -69.66\% &
  0.291 &
  1.338 \\
 &
  \amazonbaselinetable &
  \ding{55} &
  0.641 &
  18.70\% &
  0.779 &
  10.97\% &
  0.084 &
  1.087 \\
\multirow{-5}{*}{\textbf{\begin{tabular}[c]{@{}c@{}}GPT-Neo\\ (2.7B)\end{tabular}}} &
  {\color[HTML]{333333} \textbf{Ours}} &
  \ding{51} &
  {\color[HTML]{343434} \textbf{0.702}} &
  \textbf{30.00\%} &
  {\color[HTML]{343434} \textbf{0.820}} &
  \textbf{16.83\%} &
  {\color[HTML]{343434} \textbf{0.075}} &
  {\color[HTML]{343434} \textbf{1.077}} \\ \hline
\end{tabular}
}
\end{table*}

Note that we utilize the identity blocks from a different perspective than LLAMA PRO. LLAMA PRO incorporated extra multiple identity blocks into the pretrained LLM to expand the model for post-pretraining without changing the original mapping of the pretrained LLM at the initial stage, while in our method, we initialize the transformer blocks within the generator $g_{\bm{\omega}}$ as identity blocks to achieve the identity mapping of the input, thus aligning the latent space of the dynamic soft prompt with that of the target LLM $f_{\bm{\theta}}$ initially. 
\section{Experiments}
\label{sec:experiment}

\subsection{Experimental Setup}
\textbf{Models}. We evaluate~\framework~on three suites of pretrained LLMs with various scales: GPT-Neo (125M, 1.3B, 2.7B)~\citep{black_2021_5297715}, Pythia (410M, 1.4B, 2.8B, 6.9B)~\citep{biderman2023pythia} and StarCoderBase (1B, 3B, 7B)~\citep{li2023starcoder}. Both GPT-Neo and Pythia are pretrained on the Pile dataset~\citep{gao2020pile} for text generation. StarCoderBase is pretrained on The Stack dataset~\citep{kocetkov2022stack} with more than 80 programming languages for code generation.\\
\textbf{Dataset}. We extract training data of GPT-Neo and Pythia with the Language Model Extraction Benchmark dataset~\citep{memdataset}, a subset in English with 15K sequences sampled from the Pile dataset. For StarCoderBase, we utilize \textit{the-stack-smol} dataset~\cite{stacksmol}, a subset randomly sampled from The Stack dataset. In our experiments, we focus on the \textit{java}, \textit{c\#}, \textit{go} and \textit{sql} splits of it. And there are 40K sequences in total.\\
\textbf{Baselines}. We compare~\framework~with four baselines. The baseline \textit{No Prompt} corresponds to the method shown in Figure~\ref{fig:motivation} (a), serving as the vanilla baseline. We include two naive baselines by prepending hard prompt to the prefix for extraction: \textit{Constant Hard Prompt} and \textit{Dynamic Hard Prompt}. Assuming the length of the prompt is $N$, for \textit{Constant Hard Prompt}, we pick the first $N$ tokens in the vocabulary of the target LLM to serve as the hard prompt. For \textit{Dynamic Hard Prompt}, we apply the mapping $m(\cdot)$ in Section~\ref{subsec:mapping} to the prefix to generate the hard prompt without feeding it to a generator for further processing. \textit{\amazonbaseline} corresponds to the method shown in Figure~\ref{fig:motivation} (b), which is the SOTA work in the measurement of the memorization.\\
\textbf{Evaluation Settings}. The length of the prompt, prefix, and suffix is 50 by default without explicit explanation for evaluation. We use the \textit{Exact Extraction Rate (ER)}, \textit{Fractional Extraction Rate (ER)}, \textit{Test loss} and \textit{Test perplexity (PPL)} to evaluate the performance of~\framework. Note that \textit{Exact ER} corresponds to \textit{discoverable memorization rate} to estimate the verbatim memorization.

\begin{table*}[!htb]
\small
\centering
\caption{Main Results on Pythia Suite}
\label{tab:main-pythia}
\resizebox{\textwidth}{!}{
\begin{tabular}{ccccccccc}
\hline
Model &
  Method &
  \begin{tabular}[c]{@{}c@{}}Dynamic\\ Prompt?\end{tabular} &
  \begin{tabular}[c]{@{}c@{}}Exact\\ ER\end{tabular} &
  \begin{tabular}[c]{@{}c@{}}Exact ER\\ Gain\end{tabular} &
  \begin{tabular}[c]{@{}c@{}}Fractional\\ ER\end{tabular} &
  \begin{tabular}[c]{@{}c@{}}Fractional ER\\ Gain\end{tabular} &
  Test Loss &
  \begin{tabular}[c]{@{}c@{}}Test Perplexity\\ (PPL)\end{tabular} \\ \hline
 &
  No Prompt &
  N/A &
  0.236 &
  N/A &
  0.458 &
  N/A &
  0.473 &
  1.605 \\
 &
  Constant Hard Prompt &
  \ding{55} &
  0.161 &
  -31.78\% &
  0.361 &
  -21.18\% &
  0.595 &
  1.812 \\
 &
  Dynamic Hard Prompt &
  \ding{51} &
  0.039 &
  -83.47\% &
  0.119 &
  -74.02\% &
  0.704 &
  2.022 \\
 &
  \amazonbaselinetable &
  \ding{55} &
  {\color[HTML]{343434} 0.318} &
  34.75\% &
  {\color[HTML]{343434} 0.526} &
  14.90\% &
  {\color[HTML]{343434} 0.392} &
  {\color[HTML]{343434} 1.48} \\
\multirow{-5}{*}{\textbf{\begin{tabular}[c]{@{}c@{}}Pythia \\ (410M)\end{tabular}}} &
  {\color[HTML]{333333} \textbf{Ours}} &
  \ding{51} &
  {\color[HTML]{343434} \textbf{0.513}} &
  \textbf{117.37\%} &
  {\color[HTML]{343434} \textbf{0.683}} &
  \textbf{49.09\%} &
  {\color[HTML]{343434} \textbf{0.283}} &
  {\color[HTML]{343434} \textbf{1.328}} \\ \hline
 &
  {\color[HTML]{333333} No Prompt} &
  N/A &
  {\color[HTML]{333333} 0.416} &
  N/A &
  {\color[HTML]{333333} 0.648} &
  N/A &
  {\color[HTML]{333333} 0.199} &
  {\color[HTML]{333333} 1.22} \\
 &
  {\color[HTML]{333333} Constant Hard Prompt} &
  \ding{55} &
  {\color[HTML]{333333} 0.293} &
  -29.57\% &
  {\color[HTML]{333333} 0.526} &
  -18.83\% &
  {\color[HTML]{333333} 0.288} &
  {\color[HTML]{333333} 1.333} \\
 &
  {\color[HTML]{333333} Dynamic Hard Prompt} &
  \ding{51} &
  {\color[HTML]{333333} 0.067} &
  -83.89\% &
  {\color[HTML]{333333} 0.159} &
  -75.46\% &
  {\color[HTML]{333333} 0.412} &
  {\color[HTML]{333333} 1.51} \\
 &
  {\color[HTML]{333333} \amazonbaselinetable} &
  \ding{55} &
  {\color[HTML]{333333} 0.497} &
  19.47\% &
  {\color[HTML]{333333} 0.714} &
  10.16\% &
  {\color[HTML]{333333} 0.126} &
  {\color[HTML]{333333} 1.135} \\
\multirow{-5}{*}{\textbf{\begin{tabular}[c]{@{}c@{}}Pythia \\ (1.4B)\end{tabular}}} &
  {\color[HTML]{333333} \textbf{Ours}} &
  \ding{51} &
  {\color[HTML]{333333} \textbf{0.617}} &
  \textbf{48.32\%} &
  {\color[HTML]{333333} \textbf{0.786}} &
  \textbf{21.36\%} &
  {\color[HTML]{333333} \textbf{0.109}} &
  {\color[HTML]{333333} \textbf{1.115}} \\ \hline
 &
  {\color[HTML]{333333} No Prompt} &
  N/A &
  {\color[HTML]{000000} 0.517} &
  N/A &
  {\color[HTML]{000000} 0.735} &
  N/A &
  {\color[HTML]{000000} 0.144} &
  {\color[HTML]{000000} 1.155} \\
 &
  {\color[HTML]{333333} Constant Hard Prompt} &
  \ding{55} &
  {\color[HTML]{000000} 0.401} &
  -22.44\% &
  {\color[HTML]{000000} 0.611} &
  -16.87\% &
  {\color[HTML]{000000} 0.214} &
  {\color[HTML]{000000} 1.239} \\
 &
  {\color[HTML]{333333} Dynamic Hard Prompt} &
  \ding{51} &
  {\color[HTML]{000000} 0.091} &
  -82.40\% &
  {\color[HTML]{000000} 0.198} &
  -73.06\% &
  {\color[HTML]{000000} 0.33} &
  {\color[HTML]{000000} 1.39} \\
 &
  {\color[HTML]{333333} \amazonbaselinetable} &
  \ding{55} &
  {\color[HTML]{000000} 0.585} &
  13.15\% &
  {\color[HTML]{000000} 0.783} &
  6.57\% &
  {\color[HTML]{000000} 0.090} &
  {\color[HTML]{000000} 1.094} \\
\multirow{-5}{*}{\textbf{\begin{tabular}[c]{@{}c@{}}Pythia \\ (2.8B)\end{tabular}}} &
  {\color[HTML]{333333} \textbf{Ours}} &
  \ding{51} &
  {\color[HTML]{000000} \textbf{0.669}} &
  \textbf{29.40\%} &
  {\color[HTML]{000000} \textbf{0.827}} &
  \textbf{12.57\%} &
  {\color[HTML]{000000} \textbf{0.080}} &
  {\color[HTML]{000000} \textbf{1.084}} \\ \hline
 &
  {\color[HTML]{333333} No Prompt} &
  N/A &
  {\color[HTML]{000000} 0.561} &
  N/A &
  {\color[HTML]{000000} 0.781} &
  N/A &
  {\color[HTML]{000000} 0.104} &
  {\color[HTML]{000000} 1.11} \\
 &
  {\color[HTML]{333333} Constant Hard Prompt} &
  \ding{55} &
  {\color[HTML]{000000} 0.446} &
  -20.50\% &
  {\color[HTML]{000000} 0.674} &
  -13.70\% &
  {\color[HTML]{000000} 0.165} &
  {\color[HTML]{000000} 1.179} \\
 &
  {\color[HTML]{333333} Dynamic Hard Prompt} &
  \ding{51} &
  {\color[HTML]{000000} 0.122} &
  -78.25\% &
  {\color[HTML]{000000} 0.231} &
  -70.42\% &
  {\color[HTML]{000000} 0.262} &
  {\color[HTML]{000000} 1.3} \\
 &
  {\color[HTML]{333333} \amazonbaselinetable} &
  \ding{55} &
  {\color[HTML]{000000} 0.648} &
  16.04\% &
  {\color[HTML]{000000} 0.831} &
  6.67\% &
  {\color[HTML]{000000} 0.063} &
  {\color[HTML]{000000} 1.065} \\
\multirow{-5}{*}{\textbf{\begin{tabular}[c]{@{}c@{}}Pythia \\ (6.9B)\end{tabular}}} &
  {\color[HTML]{333333} \textbf{Ours}} &
  \ding{51} &
  {\color[HTML]{000000} \textbf{0.702}} &
  \textbf{25.13\%} &
  {\color[HTML]{000000} \textbf{0.858}} &
  \textbf{9.89\%} &
  {\color[HTML]{000000} \textbf{0.062}} &
  {\color[HTML]{000000} \textbf{1.064}} \\ \hline
\end{tabular}
}
\end{table*}
\begin{table*}[!htb]
\small
\centering
\caption{Main Results on StarCoderBase Suite
}
\label{tab:main-starcoderbase}
\resizebox{\textwidth}{!}{\begin{tabular}{ccccccccc}
\hline
Model &
  Method &
  \begin{tabular}[c]{@{}c@{}}Dynamic\\ Prompt?\end{tabular} &
  \begin{tabular}[c]{@{}c@{}}Exact\\ ER\end{tabular} &
  \begin{tabular}[c]{@{}c@{}}Exact ER\\ Gain\end{tabular} &
  \begin{tabular}[c]{@{}c@{}}Fractional\\ ER\end{tabular} &
  \begin{tabular}[c]{@{}c@{}}Fractional ER\\ Gain\end{tabular} &
  Test Loss &
  \begin{tabular}[c]{@{}c@{}}Test Perplexity\\ (PPL)\end{tabular} \\ \hline
 &
  No Prompt &
  N/A &
  0.062 &
  N/A &
  0.232 &
  N/A &
  0.836 &
  2.306 \\
 &
  Constant Hard Prompt &
  \ding{55} &
  0.035 &
  -43.55\% &
  0.206 &
  -11.21\% &
  0.959 &
  2.608 \\
 &
  Dynamic Hard Prompt &
  \ding{51} &
  0.006 &
  -90.32\% &
  0.066 &
  -71.55\% &
  0.958 &
  2.605 \\
 &
  \amazonbaselinetable &
  \ding{55} &
  0.071 &
  14.52\% &
  0.235 &
  1.29\% &
  0.815 &
  2.259 \\
\multirow{-5}{*}{\textbf{\begin{tabular}[c]{@{}c@{}}StarCoderBase\\ (1B)\end{tabular}}} &
  {\color[HTML]{333333} \textbf{Ours}} &
  \ding{51} &
  \textbf{0.082} &
  \textbf{32.26\%} &
  \textbf{0.244} &
  \textbf{5.17\%} &
  \textbf{0.806} &
  \textbf{2.238} \\ \hline
 &
  No Prompt &
  N/A &
  0.071 &
  N/A &
  0.254 &
  N/A &
  0.745 &
  2.106 \\
 &
  Constant Hard Prompt &
  \ding{55} &
  0.043 &
  -39.44\% &
  0.232 &
  -8.66\% &
  0.834 &
  2.302 \\
 &
  Dynamic Hard Prompt &
  \ding{51} &
  0.018 &
  -74.65\% &
  0.093 &
  -63.39\% &
  0.838 &
  2.312 \\
 &
  \amazonbaselinetable &
  \ding{55} &
  0.081 &
  14.08\% &
  0.249 &
  -1.97\% &
  0.734 &
  2.084 \\
\multirow{-5}{*}{\textbf{\begin{tabular}[c]{@{}c@{}}StarCoderBase\\ (3B)\end{tabular}}} &
  {\color[HTML]{333333} \textbf{Ours}} &
  \ding{51} &
  \textbf{0.094} &
  \textbf{32.39\%} &
  \textbf{0.268} &
  \textbf{5.51\%} &
  \textbf{0.713} &
  \textbf{2.039} \\ \hline
 &
  No Prompt &
  N/A &
  0.091 &
  N/A &
  0.277 &
  N/A &
  0.67 &
  1.954 \\
 &
  Constant Hard Prompt &
  \ding{55} &
  0.021 &
  -76.92\% &
  0.243 &
  -12.27\% &
  0.765 &
  2.149 \\
 &
  Dynamic Hard Prompt &
  \ding{51} &
  0.037 &
  -59.34\% &
  0.137 &
  -50.54\% &
  0.744 &
  2.104 \\
 &
  \amazonbaselinetable &
  \ding{55} &
  0.1 &
  9.89\% &
  0.278 &
  0.36\% &
  0.657 &
  1.928 \\
\multirow{-5}{*}{\textbf{\begin{tabular}[c]{@{}c@{}}StarCoderBase\\ (7B)\end{tabular}}} &
  {\color[HTML]{333333} \textbf{Ours}} &
  \ding{51} &
  \textbf{0.11} &
  \textbf{20.88\%} &
  \textbf{0.289} &
  \textbf{4.33\%} &
  \textbf{0.641} &
  \textbf{1.898} \\ \hline
\end{tabular}
}
\end{table*}

\subsection{Main Results}
The main results to evaluate our method and the SOTA baselines are summarized in Table~\ref{tab:main-gptneo},~\ref{tab:main-pythia} and~\ref{tab:main-starcoderbase} for the suites of GPT-Neo, Pythia and StarCoderBase, respectively.

In the application of text generation, our method can outperform all the baselines consistently and significantly. For GPT-Neo suite, compared with the vanilla baseline (i.e., \textit{No Prompt}), our method can achieve a relative improvement of 112.75\%, 41.52\% and 30.0\% in terms of \textit{Exact ER} with the model size of 125M, 1.3B and 2.7B, respectively. For the Pythia suite, our method can achieve a relative improvement of 117.37\%, 48.32\%, 29.4\% and 25.13\% over the vanilla baseline in terms of \textit{Exact ER} with the model size of 410M, 1.4B, 2.8B and 6.9B, respectively.

In the application of code generation, our method can also outperform all the baselines consistently and significantly. For StarCodeBase suite, our method can achieve a relative improvement of 32.26\%, 32.39\% and 20.88\% over the vanilla baseline in terms of \textit{Exact ER} with the model size of 1B, 3B, and 7B, respectively.

\begin{figure*}[!htb]
	\centering
	\subfloat{
	\begin{minipage}[b]{.23\linewidth}
			\centering
			\includegraphics[width=.99\textwidth]
            {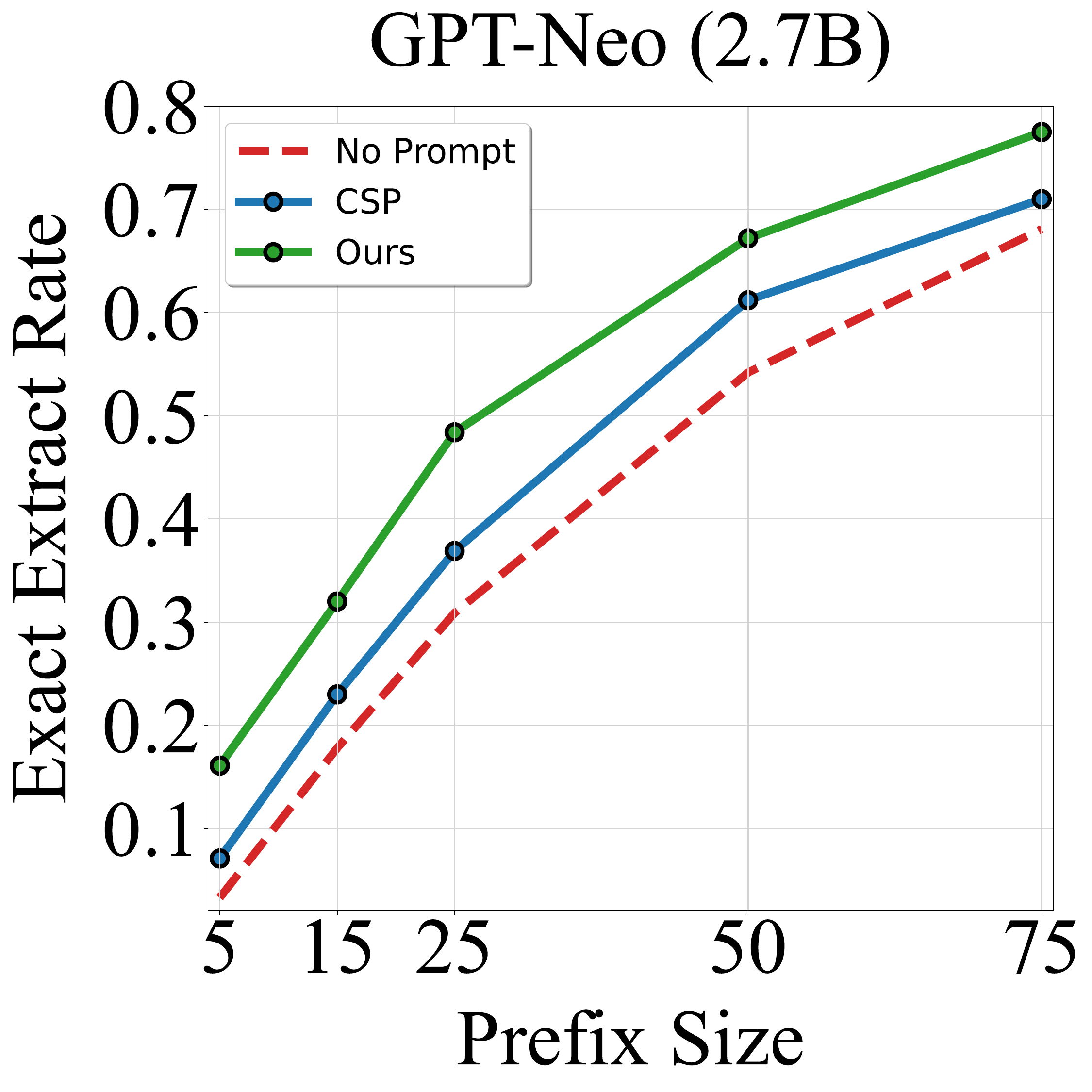}
	\end{minipage}~\label{ablation-l}}
	\subfloat{
		\begin{minipage}[b]{.23\linewidth}
			\centering
			\includegraphics[width=.99\textwidth]{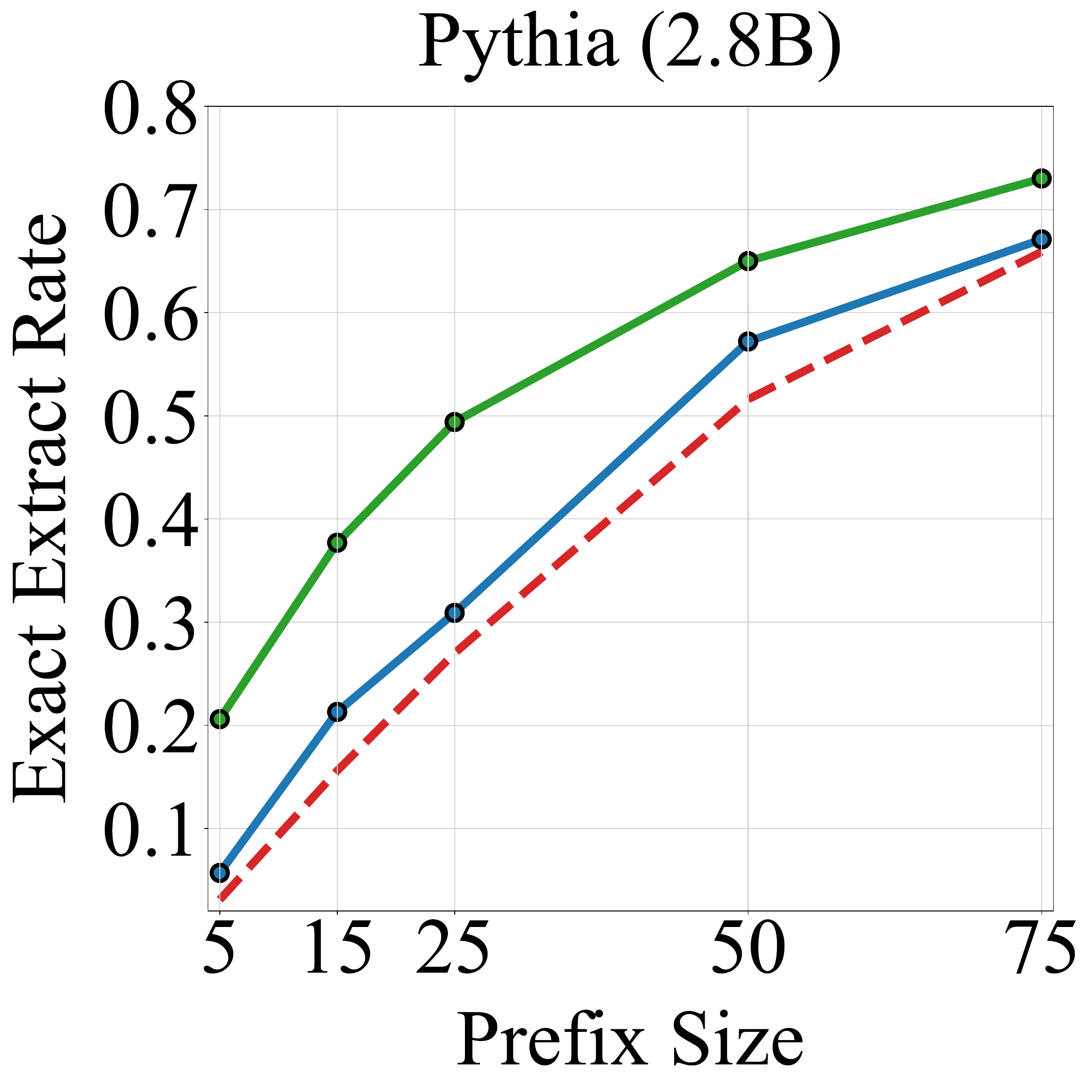}
	\end{minipage}~\label{ablation-r}}
	\subfloat{
		\begin{minipage}[b]{.23\linewidth}
			\centering
			\includegraphics[width=.99\textwidth]{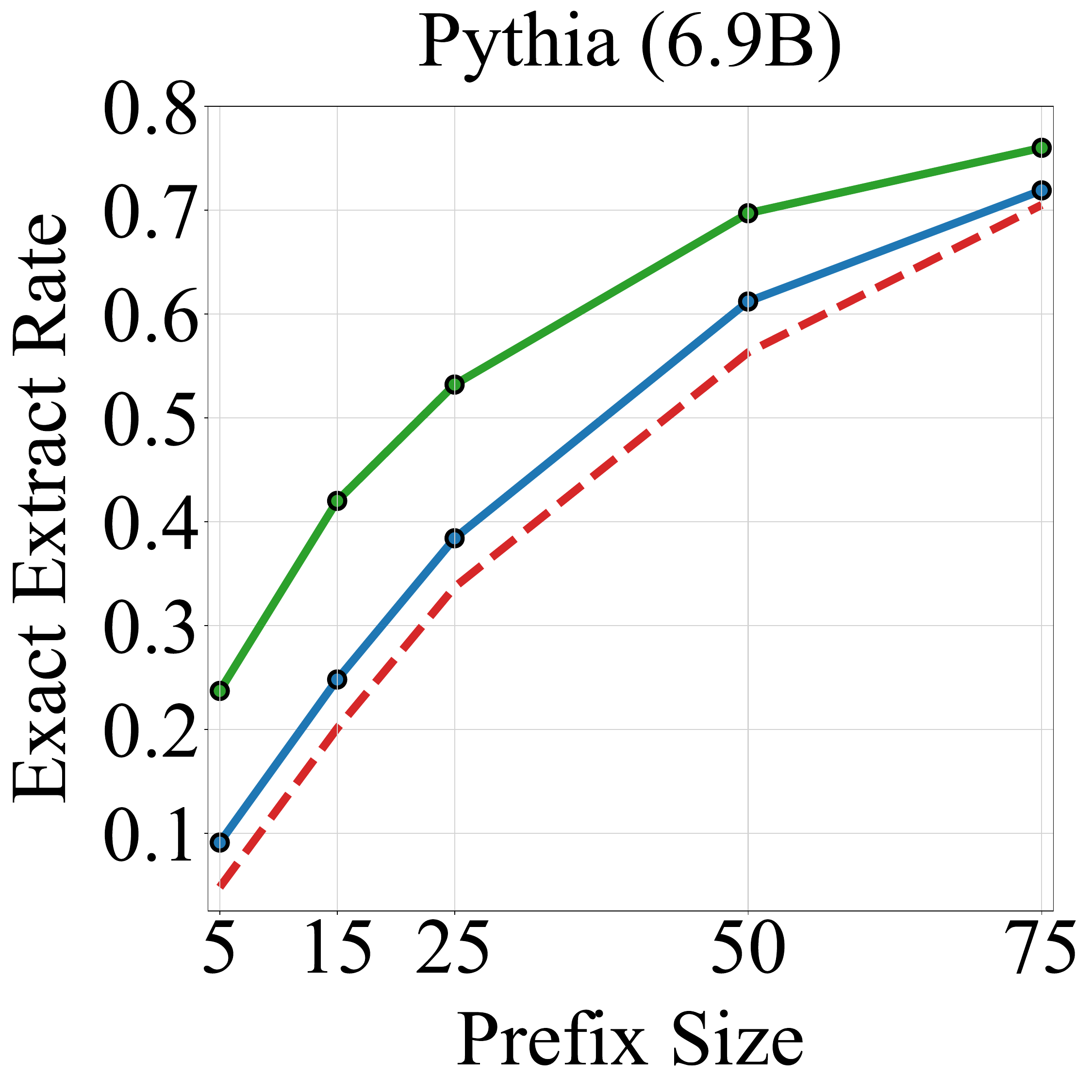}
	\end{minipage}~\label{ablation-p}}
	\subfloat{
		\begin{minipage}[b]{.23\linewidth}
			\centering
			\includegraphics[width=.99\textwidth]{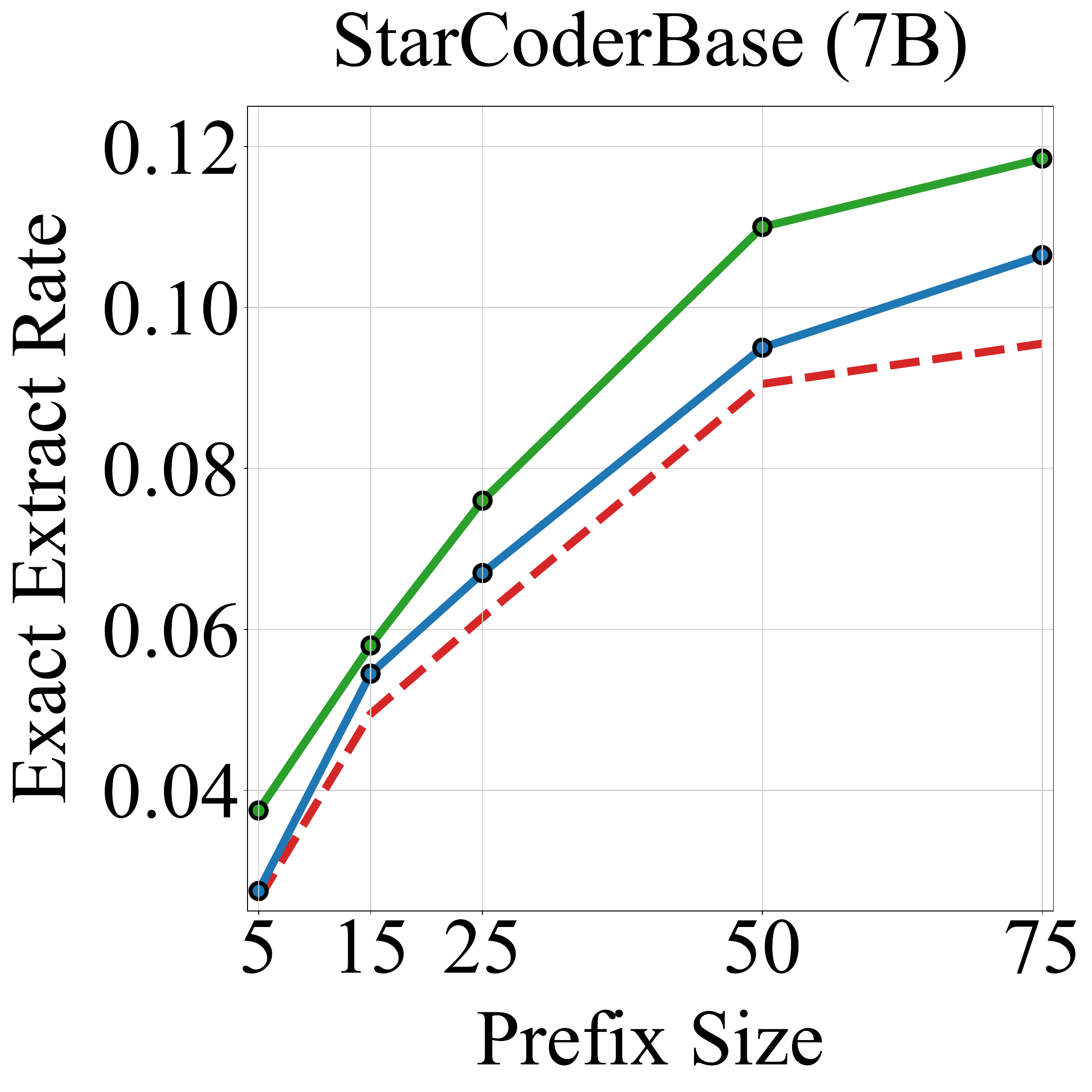}
	\end{minipage}~\label{ablation-iterations}}\\[10pt]
        	\centering
	\subfloat{
	\begin{minipage}[b]{.23\linewidth}
			\centering
			\includegraphics[width=.99\textwidth]{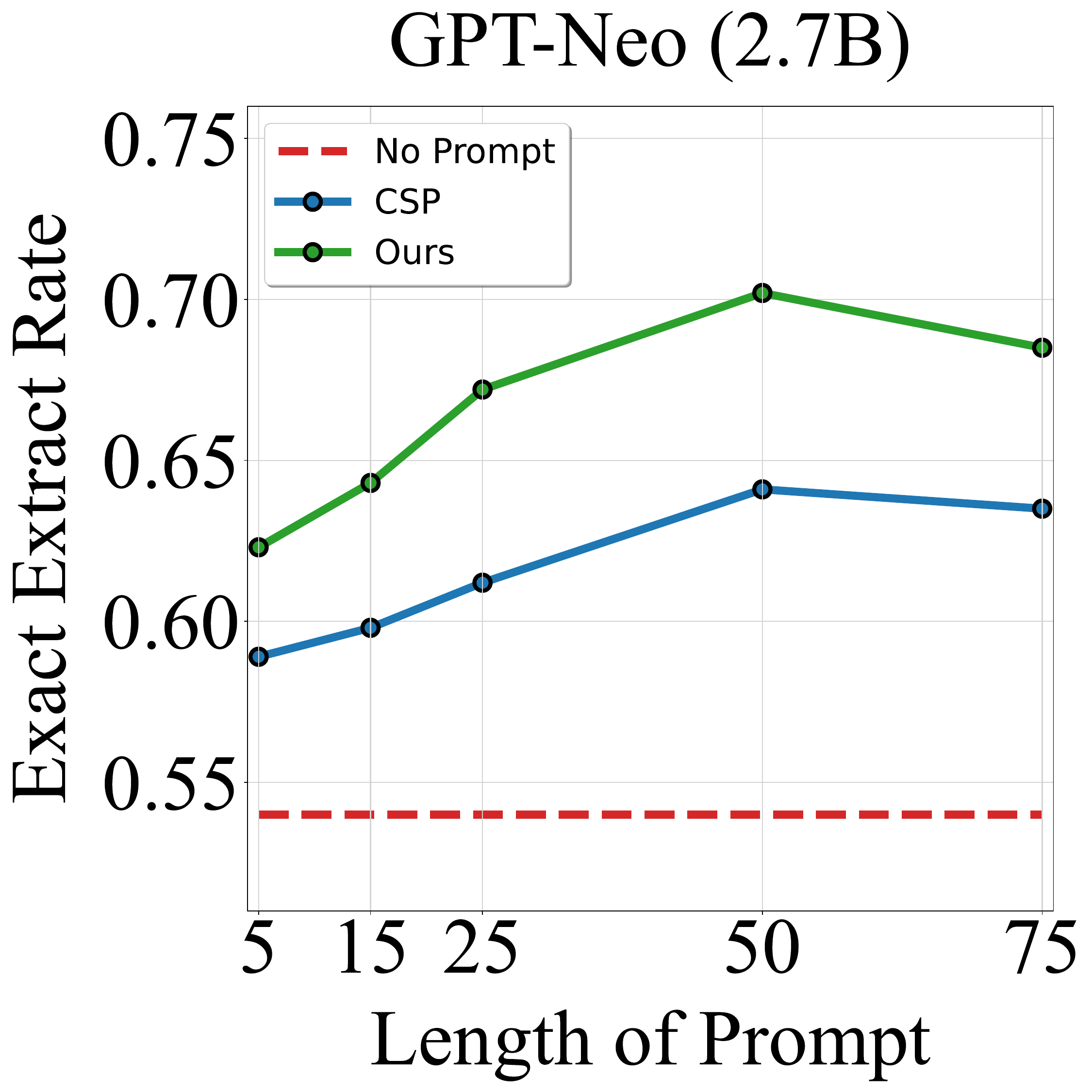}
	\end{minipage}~\label{p-l}}
	\subfloat{
		\begin{minipage}[b]{.23\linewidth}
			\centering
			\includegraphics[width=.99\textwidth]{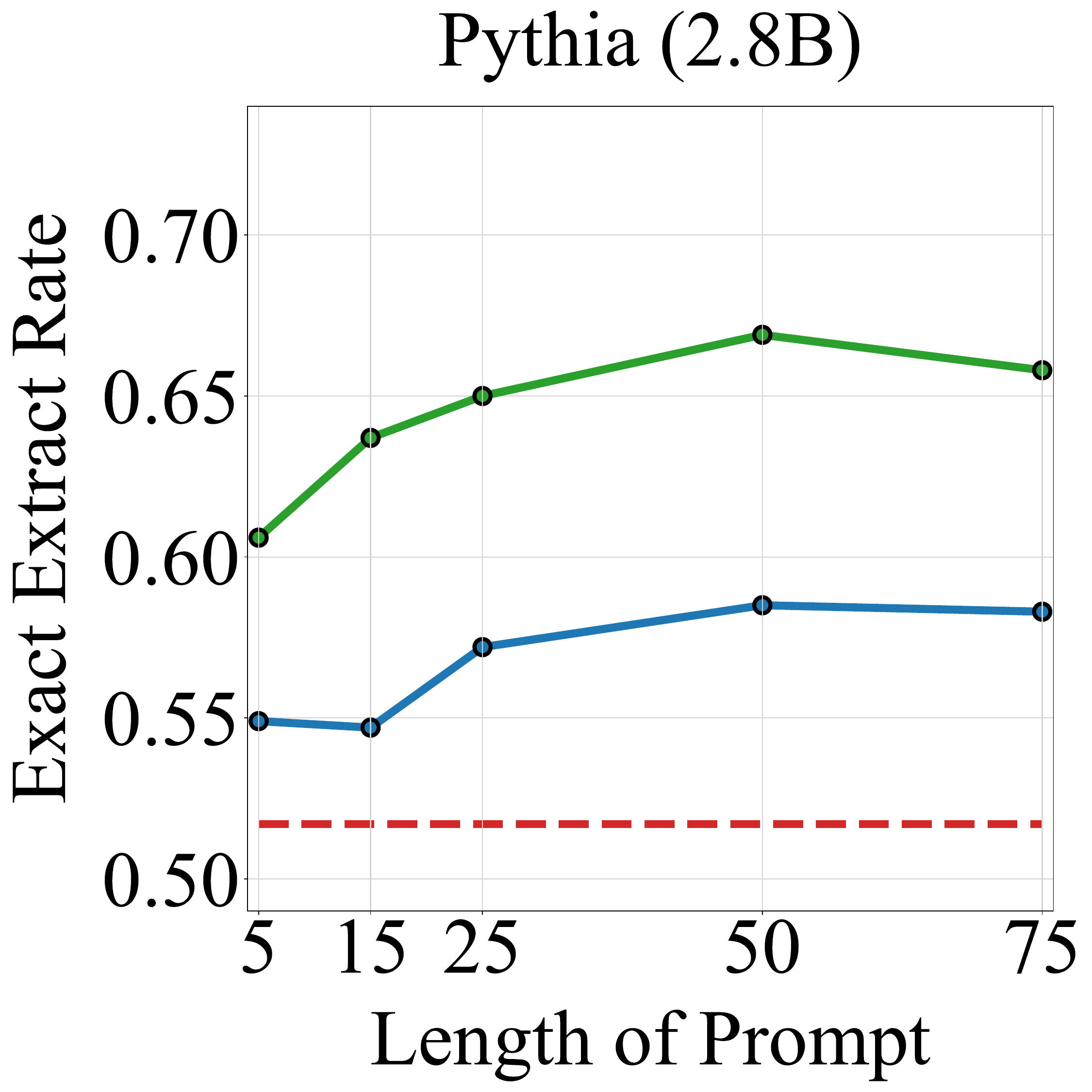}
	\end{minipage}~\label{p-r}}
	\subfloat{
		\begin{minipage}[b]{.23\linewidth}
			\centering
			\includegraphics[width=.99\textwidth]{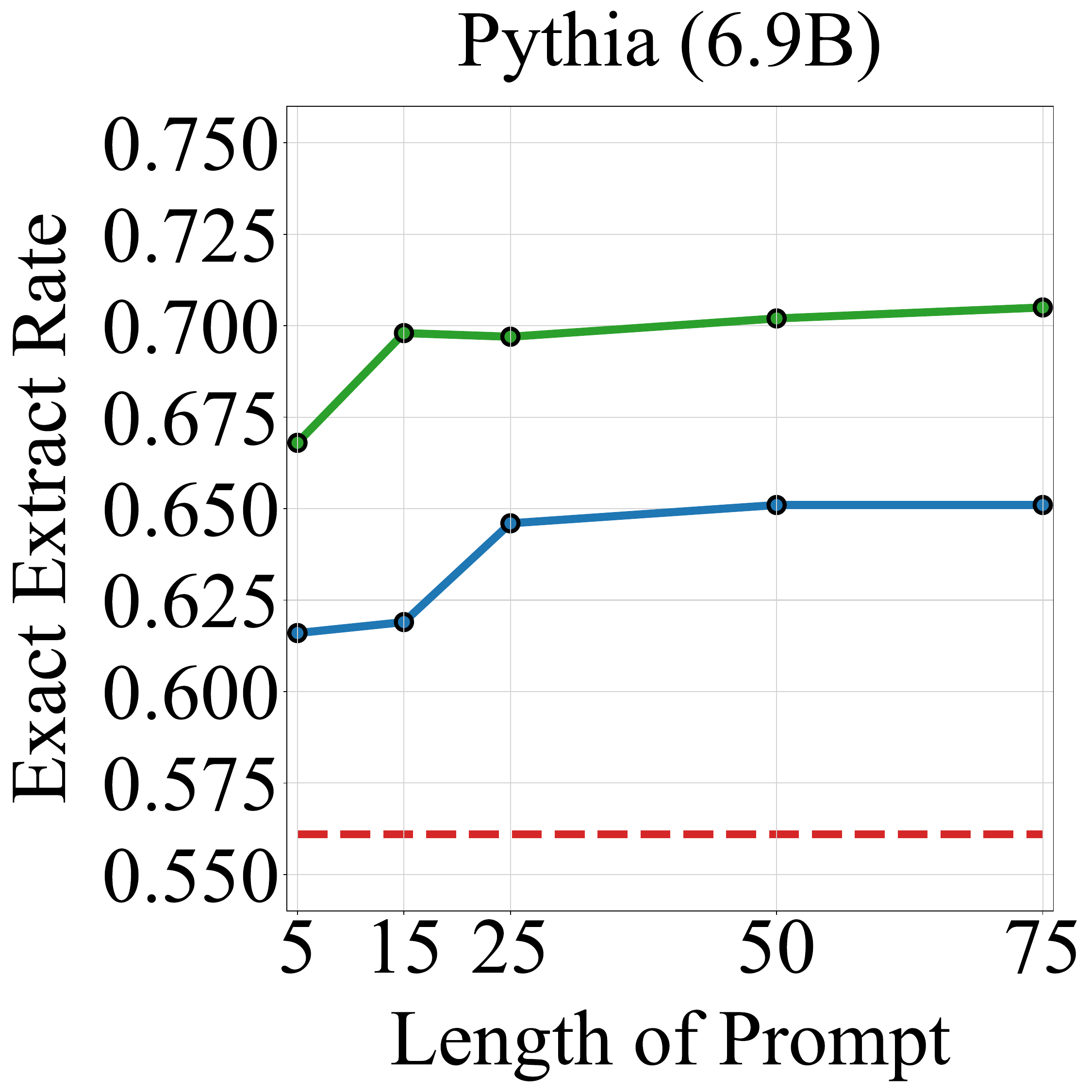}
	\end{minipage}~\label{throughput}}
	\subfloat{
		\begin{minipage}[b]{.23\linewidth}
			\centering
			\includegraphics[width=.99\textwidth]{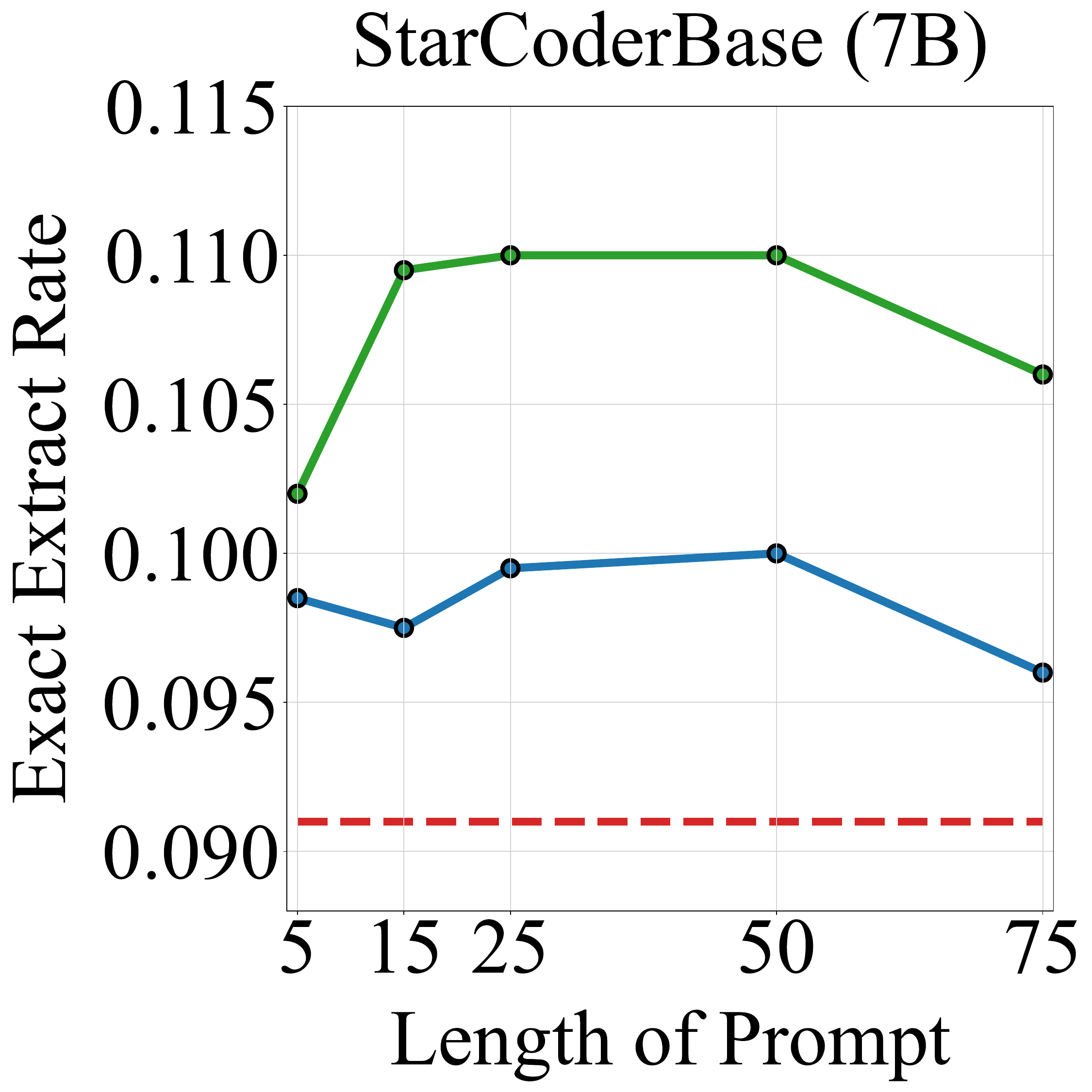}
	\end{minipage}~\label{costs}}
 
	\caption{Ablation Study on Prefix Size and Length of Prompt with Exact Extraction Rate (ER)}
	\label{fig:abs}
\end{figure*}

We have several observations from the main results across diverse settings. Firstly, prepending naive hard prompts such as \textit{Constant Hard Prompt} and \textit{Dynamic Hard Prompt} is not useful but harmful for the exaction of training data from target LLM, leading to a much lower estimation of its memorization. Secondly, our method outperforms the SOTA work, \textit{\amazonbaseline}, across the board, highlighting the importance of dynamic soft prompts for the measure of memorization. Moreover, the memorization of LLM increases with the model size, which is consistent with the existing works~\citep{carlini2023quantifying, ozdayi2023controlling, nasr2023scalable}. However, it does not mean that the small model does not have security concerns on memorization. As shown in Table~\ref{tab:main-gptneo} and Table~\ref{tab:main-pythia}, with our method, the memorization of small language models with millions of parameters is underestimated by a large margin, where their memorization cannot be ignored in the real applications.

\subsection{Ablation Study}
\label{subsec:abs}
We explore our methods from several perspectives: the impact of dynamic prompt, prefix size $L$, and the length of prompt $N$.

\begin{table}[!t]
\centering
\caption{Ablation Study on the Dynamics of Prompt}
\label{tab:dynamics-input}
\resizebox{0.45\textwidth}{!}{
\begin{tabular}{ccccc}
\hline
Model &
  Method &
  \begin{tabular}[c]{@{}c@{}}Is Dynamic \\ Prompt?\end{tabular} &
  \begin{tabular}[c]{@{}c@{}}Exact \\ ER\end{tabular} &
  \begin{tabular}[c]{@{}c@{}}Fractional \\ ER\end{tabular} \\ \hline
 &
  {\color[HTML]{333333} CSP} &
  No &
  {\color[HTML]{333333} 0.641} &
  0.779 \\
 &
  Ours &
  No &
  0.630 &
  0.765 \\
\multirow{-3}{*}{\textbf{\begin{tabular}[c]{@{}c@{}}GPT-Neo \\ (2.7B)\end{tabular}}} &
  Ours &
  Yes &
  \textbf{0.702} &
  \textbf{0.820} \\ \hline
 &
  CSP &
  No &
  0.585 &
  0.783 \\
 &
  Ours &
  No &
  0.565 &
  0.766 \\
\multirow{-3}{*}{\textbf{\begin{tabular}[c]{@{}c@{}}Pythia\\ (2.8B)\end{tabular}}} &
  Ours &
  \textbf{Yes} &
  \textbf{0.669} &
  \textbf{0.827} \\ \hline
 &
  CSP &
  No &
  0.081 &
  0.249 \\
 &
  \textbf{Ours} &
  No &
  0.081 &
  0.247 \\
\multirow{-3}{*}{\textbf{\begin{tabular}[c]{@{}c@{}}StarCoderBase\\ (3B)\end{tabular}}} &
  \textbf{Ours} &
  \textbf{Yes} &
  \textbf{0.094} &
  \textbf{0.268} \\ \hline
\end{tabular}
}
\end{table}
\noindent\textbf{Impact of Dynamic Prompt.} To evaluate the impact of dynamic prompt, we build another baseline by replacing the input to the generator with the first $N$ tokens in the vocabulary of the target LLM. In this way, the soft prompts from the generator are constant and independent of the prefix tokens. The results are shown in Table~\ref{tab:dynamics-input}. It can be observed that our method with dynamic prompt outperforms the case with constant prompt consistently and significantly over all the evaluated settings. Moreover, the performance of our method with constant prompts is close to that of directly learning a constant soft prompt. Therefore, we can conclude that the advantage of our method over \textit{CSP} comes from its adaptation to the dynamics of input instead of incorporating a transformer-based generator straightforwardly.

\noindent\textbf{Impact of Prefix Size.} To evaluate the impact prefix size, we set the length of prompt $N$ to 25 and vary the prefix size for GPT-Neo (2.7B), Pythia (2.8B), Pythia (6.9B) and StarCoderBase (7B). The results in terms of \textit{Exact ER} are shown in the first row of Figure~\ref{fig:abs}. Our method can outperform the two representative baselines consistently over all the settings of prefix size across diverse LLMs and datasets. Moreover, the amount of extracted data increases along with the increase in the prefix size, consistent with the existing works~\citep{carlini2023quantifying, ozdayi2023controlling}. 

\noindent\textbf{Impact of Length of Prompt.} To evaluate the impact length of prompt $N$, we set the prefix size $L$ to 50 and vary the length of prompt $N$ for GPT-Neo (2.7B), Pythia (2.8B), Pythia (6.9B) and StarCoderBase (7B). The results in terms of \textit{Exact ER} are shown in the second row of Figure~\ref{fig:abs}. According to the results, we have several observations. Firstly, our method can outperform the two representative baselines consistently over all the settings of length of prompt $N$ across diverse LLMs and datasets. Secondly, the performance of our method usually increases rapidly when increasing the length of prompt $N$ from a small value (e.g., 5) to a moderate value (e.g., 25). Then the performance improvement usually becomes smaller when the length of prompt $N$ is increased further. And it tends to saturate when the length of prompt $N$ reaches a relatively large value (e.g., 50 or 75). 




\section{Conclusion}
We propose a novel method to unlock memorization in large language models (LLMs) which was underestimated by previous methods. More specifically, a transformer-based generator is developed to customize the dynamic, prefix-dependent soft prompts to measure the LLM memorization. It can have a more precise detection of memorized data, capturing the data omitted by the previous methods only relying on the prefixes or the concatenation of a constant soft prompt and prefixes. Extensive experiments are conducted to show that our method can outperform the state-of-the-art techniques by a large margin under diverse settings, including text generation and code generation tasks.

\section{Limitations}

There are several limitations of our work. First, we primarily focus on the memorization of pretrained LLM over the pretraining dataset and show that our method can extract more training data. However, it has been shown that fine-tuned LLMs also have memorization on fine-tuning dataset~\citep{zeng2023exploring}. Therefore, the effectiveness of our method under the fine-tuning settings remains unexplored, including fine-tuning on a single task and multiple tasks. Second, we observed the saturation phenomenon in the ablation study on the length of prompt. The reason for the saturation remains unknown. And further studies on saturation might help extract more data with our method and thus provide better measurement of memorization. Third, based on the experimental results, we can observe that the improvement of our method on the fractional extraction rate is smaller and less robust compared with the improvement in the exact extraction rate. One possible reason is that the aligned CLM loss to train the generator is more suitable for the optimization of verbatim memorization. Since fractional extraction rate may be more important in cases where the meaning of the extracted sequences is more important than the exact match, it is valuable to improve the performance of our method on the metric of fractional extraction rate.

\section{Ethical Considerations}
In this work, we propose to leverage dynamic soft prompts to extract more training data from the target LLM and measure its memorization under the white-box settings. Therefore, it is possible that the attackers might utilize our method to extract sensitive data from the target LLM if they have white-box access to the target LLM. However, the main purpose of this work is to raise awareness among LLM researchers and developers about the security concerns caused by LLM memorization. By utilizing our method to evaluate the memorization of the target LLM, the owner of the LLM can evaluate its security vulnerability more accurately and thoroughly and then take action to defend against it. For example, we mentioned in the paper that the developer can utilize machine unlearning to forget the sensitive training data that is identified by our method. To minimize the security issues caused by our work, all of our experiments are conducted on public datasets that have been extensively studied by the research community.

\bibliography{custom, new_add, memorization, arxiv_ref}

\end{document}